\documentclass[authoryear,preprint,review,12pt]{elsarticle}

\usepackage{amssymb}
\usepackage{color}
\usepackage{graphicx}
\usepackage{booktabs}
\usepackage{multicol}
\usepackage{multirow}
\usepackage{amsmath}
\usepackage{comment}
\usepackage{moresize}
\usepackage{url}
\usepackage{lineno}
\usepackage{ifdraft}
\usepackage{pbox}
\usepackage{ifthen}
\usepackage{subfig}
\usepackage{xspace}
\usepackage{verbatim}
\usepackage{times}
\usepackage{latexsym}
\usepackage{xcolor}
\usepackage{fancyvrb}
\usepackage{microtype}
\usepackage{algorithmic}
\usepackage{lineno}
\usepackage[ruled,vlined]{algorithm2e}
\usepackage{arydshln}
\usepackage{tabularx}
\usepackage{rotating}
\usepackage{caption}
\usepackage{todonotes}

\DefineVerbatimEnvironment{code}{Verbatim}{fontsize=\small}

\journal{conference} 

\begin{document}

\begin{frontmatter}



\title
      {Five Psycholinguistic Characteristics for Better Interaction with Users}
\author[sym]{Sanja \v{S}tajner}
\ead{sanja.stajner@symanto.com}
\address[sym]{Symanto Research, N\"{u}rnberg, Germany}

\author[sym]{Seren Yenikent}
\ead{seren.yenikent@symanto.com}

\author[sym2]{Marc Franco-Salvador}
\address[sym2]{Symanto Research, Valencia, Spain}
\ead{marc.franco@symanto.com}

\newpage

\begin{abstract}
When two people pay attention to each other and are interested in what the other has to say or write, they almost instantly adapt their writing/speaking style to match the other. 
For a successful interaction with a user, chatbots and dialogue systems should be able to do the same.
We propose a framework consisting of five psycholinguistic textual characteristics for better human-computer interaction. 
We describe the annotation processes used for collecting the data, and benchmark five binary classification tasks, experimenting with different training sizes and model architectures. 
The best architectures noticeably outperform several baselines and achieve macro-averaged F$_1$-scores between 72\% and 96\% depending on the language and the task. 
The proposed framework proved to be fairly easy to model for various languages even with small amount of manually annotated data if right architectures are used. 
\end{abstract}

\begin{keyword}
natural language processing \sep machine learning \sep deep learning \sep psycholinguistics \sep emotionality \sep communication styles
\end{keyword}

\end{frontmatter}


\section{Introduction}
\label{sec:intro}

Communication represents the manifestation of personality which showcases one's psychological state. 
In human-to-human communication (either written or oral), people instinctively attempt to be in synchronization with each other to the extent that they want to commit to the interaction, engaging themselves in a so-called conversational dance \citep{Pennebaker2011}. 
In order to stay engaged and show their interest, parties invest efforts to switch their attention and adjust to the communication style of the other party. 
Due to less information processing necessary for adjustment, people prefer to interact with those who have better-resonating personality and communication styles \citep{WuEtAl'2017}.

The preference for more resonating personalities and communication styles extends beyond human-to-human interaction.
Tailoring advertising messages to different Big 5 personality types \citep{Costa} has been shown to increase the click and conversion rates \citep{matz2017psychological}, confirming that people prefer to engage with written materials that better resonate with their personality. 
Personality-tailored marketing materials were also more successful at persuading consumers to eat healthier, or purchase things they really need \citep{matz2017using}.

In the last few years, many studies attempted at building personalised dialogue agents, emotionally intelligent virtual agents, empathetic chatbots and dialogue systems \citep{empathyVirtualAgentsSurvey,rashkin-etal-2019-towards,EmpatheticChatbotFung19,HappyBotFung19,EmpatheticListenerFung19,ma2020survey}. 
It has been shown that adapting the style of the answers to the emotional state of a user leads to longer and more satisfactory user engagement not only in general conversation (i.e. chitchat) chatbots, but also in goal-oriented chatbots, as well as in human-to-human conversations in various tasks and domains \citep{rashkin-etal-2019-towards}. 
Some studies suggested that for building emotionally intelligent virtual agents, it is not enough to adapt them only to the emotion of the user, but also to the user's personality
\citep{ma2020survey,ma2019exploring,lee2019expressing}. 
It has been shown that adding personality traits to virtual agents leads to significantly better perceived emotional intelligence of such systems \citep{ma2019exploring}. 
From the computational perspective, however, automatically detecting someone's personality is difficult, and state-of-the-art performances are far away from those needed for real-world applications \citep{gjurkovic2018reddit,vstajner2020survey,gjurkovic2020pandora}.
The best models rarely surpass the majority-class baselines in a binary setting, even in cases where large amounts of textual data per user are available \citep{Plank'2015,Verhoeven'2016,Big5vsMBTI18}.

Building fully empathetic virtual agents has been proven to be very demanding as both approaches, theoretically-based and empirically-based, have some major shortcomings.
Theoretically-based approaches suffer from "lost in translation" problem, using theories based on behavioural data obtained from different angles, accumulated through various
studies and abstracted into computational models \citep{empathyVirtualAgentsSurvey,vstajner2021mbti}, whereas empirically-based approaches suffer from data biases, are domain dependent and not easily generalizable to other application scenarios \citep{empathyVirtualAgentsSurvey}.

We argue that for building successful task-oriented chatbots, better resonating written responses, and more engaging marketing campaigns, it is not necessary to model deep psychological characteristics, but rather the surface realizations of psychological states and user preferences expressed through five carefully selected psycholinguistic characteristics which can be modelled more precisely. 
Unlike traditional personality frameworks, which capture enduring personality types \citep{Roberts&Mroczek'2008,asendorpf2009personality}, the psycholinguistic characteristics that we propose can change instantly depending on the context, topic, and listener.
We thus find them more suitable for interactive communications across different domains and platforms.
\section{Five Psycholinguistic Characteristics}
\label{sec:method}

To design a framework that can be used on short textual utterances, we focus on emotionality and four communication styles as main aspects of instant communication that reflect psychological and contextual states of the user during an interactive communication. 
A task-oriented conversational agent should be able to comprehend the psychological state of the user signaled through those five characteristics. 
We argue that one of the major tasks for successful communication is to understand the tone of the conversation, i.e. whether the user is emotional or not, before getting into the detection of specific emotions. 
The four communication style features, in turn, consider situational elements, such as user experiences, expertise and interest in the domain, and, what is especially important for goal-oriented conversational agents, the goals and needs of the user.

\subsubsection{Emotionality}

Emotionality influences fundamental psychological experiences such as personal interests and attentiveness \citep{Krapp,Tausczik}. 
It is highly related to, and can be detected with, linguistic features \citep{Tausczik}. 
By using various linguistic features automatically extracted from text, it has been shown that people tend to use the same levels of emotionality during conversation \citep{Pennebaker2011}. 
Therefore, correctly detecting and responding to the emotional state of the user is an important aspect of an empathetic communication system.

Emotionality does not only refer to the similarity between the sender and the receiver of the message, but rather to any emotional responsiveness that takes place during the interaction \citep{Hoffman2001}. 
Emotional experiences and reactions of a person change instantly depending on the personal condition and triggers. 
Hence, the capability of an empathetic dialogue system to keep up with those dynamic states of the user and tailor the responses accordingly is fundamental for successful communication with the user. 
While emotionality is clearly a variable ranging anywhere between completely non-emotional
to highly emotional, for the annotation purposes, we define emotionality as a binary label:

\begin{itemize}
    \item {\bf Emotional}: Feeling-focused statements that focus on values and emotions;
    \item {\bf Non-emotional}: Emotionally-neutral statements that provide logical and analytical meaning.
\end{itemize}

\begin{table}[!t]
\begin{center}
\caption{Examples for emotionality aspect in communication.}\label{tab:Emotionality}
\scalebox{0.86}{
\setlength{\tabcolsep}{5pt}
\begin{tabular}{lp{12.5cm}}
\toprule
Emotionality & Example\\
\midrule
\multirow{1}{*}{Non-emotional} & I would prefer to buy this car since it is hybrid and cost effective.\\
\multirow{1}{*}{Emotional} & They offer a friendly service with great choice of drinks and stunning view!\\
\multirow{1}{*}{Mixed} & You're looking at the newest member of our team. She is ready to tear it up!
\\
\bottomrule
\end{tabular}
}

    \end{center}
\end{table}

The decision to treat emotionality as a binary label, as expected, had an impact on the observed inter-annotator agreement, leading to disagreements on instances which have a combination of emotional and non-emotional signals. 
Table~\ref{tab:Emotionality} presents the cases of clearly non-emotional and emotional instances (on which all annotators agreed), and an instance which contains a mixture of emotional and non-emotional signals (on which the annotators disagreed). 

\subsubsection{Communication Style}

An empathetic communication process includes effective stylistic associations. 
When the communication is tailored based on the style, it generates positive interactions, e.g. feeling of getting liked by others as part of the social function \citep{Bell1984}.
Our approach to communication style is based on the Four-sides Communication Model \citep{SchulzvonThun}. 
According to this model, every utterance reveals important information about the sender, the receiver, and the topic, in four different layers:

\begin{enumerate}
    \item \emph{Experience layer} which provides self-revealing information about the user;
    \item \emph{Factual layer} which contains facts and data-related information;
    \item \emph{Appeal layer} which contains the desires and effects that the user seeks; 
    \item \emph{Relationship layer} which provides indicators of how the sender feels about the receiver through intonation, body language, and gestures. 
\end{enumerate}

The model emphasizes the practical aspects of communication, and has thus been effectively used in both one-to-one contexts and mass communication strategies, in organizational learning (e.g. embedding a certain knowledge into the organizational culture) and occupational improvements (e.g. improving the communication abilities of project managers) \citep{natalia2015communication,lent2013cybernetic}.

As the model relies on the dynamic aspects of communication, it allows us to apply the layers on short text snippets which are the products of situational and domain-specific triggers, similar to those found in a communication with goal-oriented communication agents. 
To incorporate this framework into the context of human-to-human or human-computer textual
interaction, we use only the first three layers of the model, as the fourth layer (relationship layer) is based on face-to-face communication characteristics and thus not applicable to purely textual contexts. 
To satisfy the needs of goal-oriented chatbots (although our framework is also applicable in humanto-human communication), we further break down the appeal layer into two separate characteristics (action-seeking and information-seeking), to capture more granular aspects of the appeal, necessary to better respond to the user's needs.

Hence, based on the first three layers of the Four-sides Communication Model mentioned above, we define four communication styles as following:

Therefore, we define four communication styles as following:
\begin{itemize}
    \item {\bf Self-revealing}: Statements in which speaker shares personal information or experiences;
    \item {\bf Fact-oriented}: Factual and objective statements;
    \item {\bf Action-seeking}: Direct or indirect requests, suggestions, and recommendations expecting action from other people;
    \item {\bf Information-seeking}: Direct or indirect questions searching for information.
\end{itemize}

\begin{table}[!t]
\begin{center}
\caption{Examples of different communication styles.}\label{tab:Communication}
\scalebox{0.86}{
\begin{tabular}{ll}
\toprule
Communication style & Example\\
\midrule
\multirow{1}{*}{Self-revealing} & My husband was also diagnosed with a lung cancer.\\
\multirow{1}{*}{Fact-oriented} & For this phone, battery lasts about 20 minutes but excellent for price.\\
\multirow{1}{*}{Action-seeking} & Try contacting the customer service, here's the link.\\
\multirow{1}{*}{Information-seeking} & I would like to know if anyone would be interested in helping.\\
\bottomrule
\end{tabular}
}
    \end{center}
\end{table}

Examples for each of the four communication styles are provided in Table~\ref{tab:Communication}.
\section{Data Collection}
\label{sec:data}

Data was collected from various sources, ranging from those that contain very short posts (Facebook, Twitter, YouTube), through Amazon reviews, to forums where multiple users interact with each other. 
In all cases, each post by one user was treated as a separate instance, and we did not allow for multiple posts from the same user from any of the sources (to avoid user bias as much as possible).

\subsection{Annotation Procedure}
\label{sec:anno_proc}

We hired three annotators, all native speakers, and provided a workshop in which we explained the theory behind each of the five psycholinguistic characteristics, and showed them a number of examples for each. 
A question and answer session followed to clarify potential misunderstandings.
After that, the annotators were provided with a small set of 100 instances and asked to annotate each instance for each of the five characteristics (emotionality, self-revealing, fact-oriented, action-seeking, and information seeking) by assigning it a \textit{yes} or \textit{no} label. 
After the test phase was finished, we conducted the second workshop to further clarify potential difficulties of the annotation task. 
After that, each annotator was given the final annotation set.

\subsection{Inter-Annotator Agreement}
\label{sec:anno_agg}

\begin{table}[!t]
\begin{center}
\caption{Percentage of instances with perfect agreement.}\label{tab:IAA}
\setlength{\tabcolsep}{4.5pt}
\begin{tabular}{lc}
\toprule
Aspect & Perfect agreement \\
\midrule
Emotionality & 53\% \\
Fact-oriented & 52\% \\
Self-revealing & 63\% \\
Action-seeking & 73\% \\
Information-seeking & 80\% \\
\bottomrule
\end{tabular}
\end{center}
\end{table}

Table~\ref{tab:IAA} shows the percentage of instances with perfect inter-annotator agreement among three annotators for English, for each psycholinguistic characteristic separately. 
To better understand the causes of disagreement, we had a closer look at the instances in which annotators disagreed on some of the psycholinguistic characteristics, and noticed two main sources of disagreements:

\begin{enumerate}
\item The disagreements on the emotionality aspect were mostly observed for longer instances which contained a mixture of emotional and non-emotional signals. 
By talking to the annotators, we discovered that there were two main causes for annotation disagreements in those cases: (a) different annotators focused on different parts of the instances; and (b) annotators had different thresholds for the number of emotional signals they deem necessary to label a certain instance as emotional.
For example, in the following post, the first part conveys an emotional expression that is followed by a rational statement that includes reasoning:
\begin{itemize}
    \item \emph{You just hit the nail on the head! They have Kante so he's not going to get much playing time.}
\end{itemize}
\item Those instances which are emotional and fact-oriented at the same time led to most disagreements on both emotionality and fact-oriented labelling tasks. 
This revealed that humans tend to associate the fact-oriented statements with being non-emotional, and non-fact-oriented with being emotional. 
Although it is more common to have statements that are both non-emotional and fact-oriented, or those that are both emotional and non-fact-oriented, statements that are emotional and fact-oriented at the same time also exist, as in the following example:
\begin{itemize}
    \item \emph{All this is a scam to push the EV market and away from diesel fuel this was all pre set up for everyone to start buying electric and not go towards "Clean Diesel".}
\end{itemize}
\end{enumerate}

A manual inspection of 300 annotated instances from the final annotation set confirmed that disagreements were not the result of low quality annotations, but rather the natural complexity and subjectivity of the task.
\section{Experiments}
\label{sec:expts}

\begin{table}[!t]
\begin{center}
\caption{Statistics of the dataset partitions.}\label{tab:data_stats}
\setlength{\tabcolsep}{4.5pt}
\begin{tabular}{lrrcc}
\toprule
\multirow{2}{*}{Aspect} & \multicolumn{2}{c}{Training} & \multicolumn{2}{c}{Test} \\
& NO & YES & NO & YES \\
\midrule
Emotionality & 6,557 & 6,538 & 496 & 504 \\
Fact-oriented & 11,047 & 2,097 & 827 & 173 \\
Self-revealing  & 2,548 & 13,162 & 160 & 840 \\
Action-seeking & 15,969 & 1,945 & 884 & 116 \\
Information-seeking & 17,259 & 3,177 & 844 & 156 \\
\bottomrule
\end{tabular}
\end{center}
\end{table}

\subsection{Datasets}
\label{sec:exp:datasets}

Given the high number of instances that are difficult to annotate even by humans (Section~\ref{sec:anno_agg}), we did not want to bring noise into the comparison of model performances among various architectures and training sizes by having a high number of "gray zone" cases in the test sets.
Therefore, we opted for the main test sets to consist only of instances with perfect inter-annotator agreement ("clear cases"). 
When it comes to training sets, we experimented with both scenarios: (1) using only the cases on which all three annotators agreed ("clear cases" only); and (2) using the same portion of data
randomly sampled from the whole annotated datasets, taking the majority label as the "gold label". 
The first scenario yielded significantly better results than the second, and therefore, we here discuss only those results and present the statistics of those datasets (Table~\ref{tab:data_stats}). 
This scenario ensured the good quality of training data (due to training dataset consisting only of "clear cases"), and easier discussion of the results knowing that trained human annotators had perfect agreement on those test sets.

We performed a stratified division of data, aiming to have 1,000 test instances in each classification task. 
The rest of the annotated data with perfect inter-annotator agreement was used as the main (largest) training datasets. 
We also tried balancing the two classes in the training datasets of each binary classification task. 
The resulting classification performances were not better than in the unbalanced scenario, and are thus not presented in this paper.

\paragraph{Additional test set.} To assess the usefulness of our models in a real-world scenario, where given utterances might not all be the "clear cases", we built an additional test set for the emotionality classification task in English by randomly choosing 500 instances on which the annotators did not agree (where the "gold label" is the majority label) and 500 instances on which all three annotators agreed, from a new set of instances from Twitter, collected one year later than the original datasets. 
In this annotation round, we additionally asked the annotators to label the instances for being easy or difficult cases. 
Each instance that was annotated as difficult by at least two annotators, obtained a "gold label" difficult, others were labelled as easy. 
This resulted in 482 instances being labelled as easy, and 518 instances being labelled as difficult. 
We use those additional labels to calculate accuracy of our best models on the test set consisting of only difficult cases (Section~\ref{sec:res_additional}), and to check how they relate to the class probabilities in our best models~\ref{sec:res_class_probs}).

\subsection{Classification Models and Evaluation Metrics}

\begin{figure}[!t]
  \centering
	\includegraphics[width=1.0\linewidth]{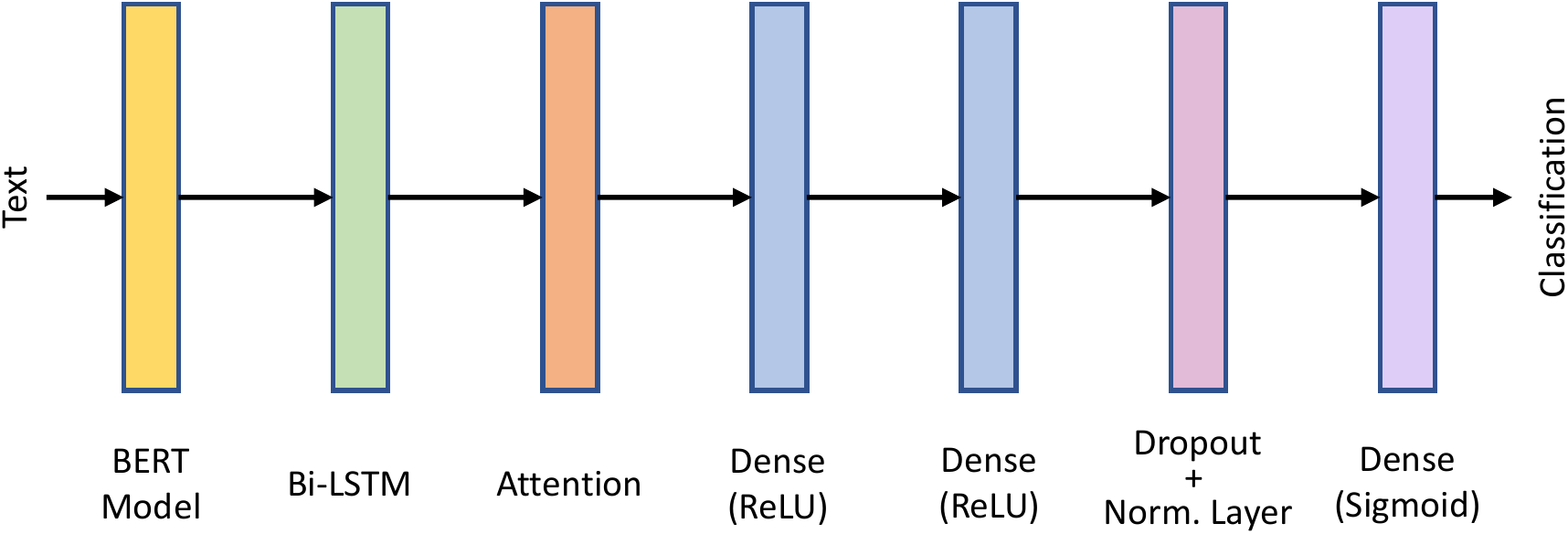}
	\caption{Classification model architecture.}
  \label{fig:1}
\end{figure} 

We use fine-tuned Bidirectional Encoder Representations from Transformers (BERT) model \citep{devlin2018bert} to obtain the word embeddings from the input text.\footnote{1For more details about BERT, its pretrained models, and how to fine-tune, please refer to: \url{https://github.com/google-research/bert}.} 
They feed a bidirectional Long Short-Term Memory (LSTM) \citep{hochreiter1997long}. 
An attention mechanism \citep{yang2016hierarchical} combines its hidden states. 
Finally, we get the classification result by using three fully connected dense layers
together with dropout and layer normalisation \citep{ba2016layer}. 
The model architecture is shown in Figure~\ref{fig:1}. 
We fine-tune a different BERT model for each training partition, using 5\% of the training data as development partition. 
We note that the BERT word representations are derived from sub-word information, which makes the model versatile in terms of vocabulary size and spelling errors.

We also explore two additional baseline models: (1) a classical word and character n-grams with Support Vector Machines \citep{chang2011libsvm} classifier (ng-SVM);\footnote{The ng-SVM parameters include the top 20k word f1,2g-grams, top 20k character \{3,4,5\}-grams with word boundaries, and the Term Frequency-Inverse Document Frequency (TF-IDF) weighting scheme \citep{salton1983extended} with minimum DF of 2.} and (2) representing the input text by means of character embeddings post-processed with Convolutional Neural Networks (c-CNN) \citep{lecun2015deep}. 
For the latter model, we use the neural architecture proposed by Zhang et al. (2015) but add a character embedding layer as input.\footnote{The c-CNN parameters include 128-dimensional trainable random embeddings, top 20k characters, maximum text length of 400, 1024 filters, and kernels of size 3, 5 and 7.}
The performance of such architecture is similar to that of the original one, but the embedding layer contributes to making the model smaller and faster converging at the training time.

\paragraph{Evaluation Metrics.} As evaluation metrics, we use the macro-averaged per-class Precision (P), Recall (R) and F1-score (F), for all models. 
We opt for the macro-averaged metrics to consider equally important each class regardless of its frequency. 
This eases the analysis of results with our imbalanced test sets.
\section{Results}
\label{sec:results}

\begin{table}
\begin{center}
\caption{Macro-averaged performances (in \%) of the compared models for all English tasks.}\label{tab:comp_results_0}
\scalebox{0.9}{
\setlength{\tabcolsep}{4.5pt}
\begin{tabular}{lccccccccccccccc}
\toprule
\multirow{2}{*}{Model} & \multicolumn{3}{c}{Emotionality} & \multicolumn{3}{c}{Fact-oriented} & \multicolumn{3}{c}{Self-revealing} & \multicolumn{3}{c}{Action-seeking} & \multicolumn{3}{c}{Info-seeking} \\
& P & R & F & P & R & F & P & R & F & P & R & F & P & R & F \\
\cmidrule(lr){1-1}
\cmidrule(lr){2-4}
\cmidrule(lr){5-7}
\cmidrule(lr){8-10}
\cmidrule(lr){11-13}
\cmidrule(lr){14-16}
c-CNN & 89 & 89 & 89 & 89 & 80 & 83 & 86 & 86 & 86 & 85 & 81 & 83 & 94 & 93 & 94 \\
ng-SVM & 90 & 90 & 90 & 90 & 87 & 88 & 89 & 86 & 87 & 91 & 81 & 85 & 94 & 88 & 91 \\
BERT & 95 & 95 & 95 & 94 & 95 & 95 & 94 & 96 & 95 & 92 & 87 & 90 & 96 & 96 & 96 \\
\bottomrule
\end{tabular}
}
\end{center}
\end{table}

The macro-averaged results of the classification experiments on all five psycholinguistic characteristics and for all three model architectures, using the full English training datasets are
presented in Table~\ref{tab:comp_results_0}. 
The BERT models perform best on all tasks, while ng-SVM performs better than c-CNN on average, showing worse performance only on the information-seeking classification task.

\subsection{Results on Additional Test Set}
\label{sec:res_additional}

To discard any kind of data artifact that could ease the classification of the main test instances, we also test our best emotionality classification model (BERT) on the additional
test set (Section \ref{sec:exp:datasets}). 
The model achieves an 81\% macro-averaged F1-score on this additional test set.
On the cases labeled as difficult, it achieves a 74\% macro-averaged F1-score. 
Considering the additional difficulty of social media texts, the topic-domain shift produced due to one year between the crawling of the training and the test sets, and inclusion of difficult cases in the test set, we consider this result as a proof of our model's satisfactory performance. 
These results also extend the validity of our best models to the real-world applications where quality and versatility to adapt to new domains is a must.

\subsection{Class Probabilities for Assessing Emotional Intensity}
\label{sec:res_class_probs}

By using the additional test set, we find that the class probability of our best emotionality model might be used to assess the emotional intensity of the post.
In over 94\% of the test instances, easy cases of non-emotional instances have a probability of the no class between 85\% and 99\%, while difficult cases have a probability of no class between 40\% and 60\%. 
These results indicate that the class probability might be used as a measure of the emotional intensity for applications where a binary label is not sufficient.

\subsection{Influence of Training Size}
\label{sec:res_influence}

\begin{figure}[!t]
  \centering
	\includegraphics[width=0.95\linewidth]{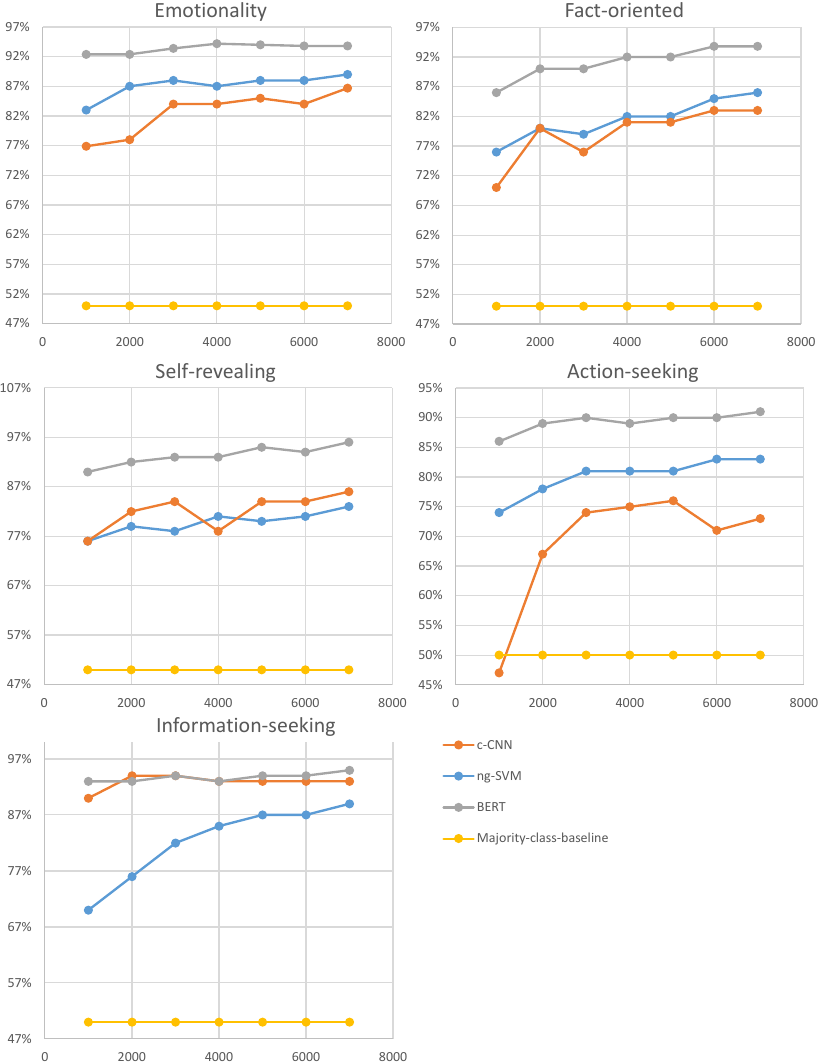}
	\caption{Macro-averaged F$_1$-score of the compared models in function of the training size.}
  \label{fig:3.1}
\end{figure} 

We also investigate the influence of training size on the performance of our best models. 
The macro-averaged F1-score obtained in those experiments is plotted in Figure~\ref{fig:3.1}. 
As can be seen, even with using only 1,000 instances for training, the BERT model achieves over 92\% macro-averaged F1-score for emotionality detection. 
The results of the binary classifiers trained on smaller portions of the training data (even only 1,000 instances) for the four communication characteristics also achieve very high performance, significantly outperforming the majority-class baseline in all cases.

\subsection{Other Languages}
\label{sec:other_langs}

\begin{table}
\begin{center}
\caption{Macro-averaged performances (in \%) for all the languages and tasks using the BERT architecture.}\label{tab:data_results_0}
\scalebox{0.9}{
\setlength{\tabcolsep}{4.5pt}
\begin{tabular}{lccccccccccccccc}
\toprule
\multirow{2}{*}{Language} & \multicolumn{3}{c}{Emotionality} & \multicolumn{3}{c}{Fact-oriented} & \multicolumn{3}{c}{Self-revealing} & \multicolumn{3}{c}{Action-seeking} & \multicolumn{3}{c}{Info-seeking} \\
& P & R & F & P & R & F & P & R & F & P & R & F & P & R & F \\
\cmidrule(lr){1-1}
\cmidrule(lr){2-4}
\cmidrule(lr){5-7}
\cmidrule(lr){8-10}
\cmidrule(lr){11-13}
\cmidrule(lr){14-16}
English & 95 & 95 & 95 & 94 & 95 & 95 & 94 & 96 & 95 & 92 & 87 & 90 & 96 & 96 & 96 \\
Spanish & 94 & 93 & 94 & 96 & 96 & 96 & 93 & 93 & 93 & 79& 76 & 77 & 97 & 96 & 96 \\
German & 86 & 86 & 86 & 88 & 88 & 88 & 93 & 92 & 93 & 88 & 84 & 86 & 95 & 92 & 94 \\
Arabic & 88 & 88 & 88 & 83 & 79 & 81 & 90 & 84 & 86 & 80 & 67 & 72 & 93 & 93 & 93 \\
Chinese & 91 & 90 & 91 & 92 & 92 & 92 & 90 & 91 & 90 & 87 & 86 & 87 & 95 & 96 & 95 \\
\bottomrule
\end{tabular}
}
\end{center}
\end{table}

To test applicability of the proposed method in languages other than English, we collected the data from the same sources and used the same annotation procedures for four other languages: Spanish, German, Arabic, and Chinese. 
All test sets were of the same size (1,000 instances) following the class distribution of their corresponding training datasets. 
Training datasets in all languages were of similar sizes. 
The macro-averaged performances of BERT models for all languages and tasks are provided in Table~\ref{tab:data_results_0}.

\subsection{Error Analysis}
\label{sec:res_err_an}

We performed a manual error analysis on all misclassified instances by the English models trained with BERT on the full training datasets.

The majority of errors in emotionality detection seem to stem from the model's inability to recognize sarcasm and some fixed expressions which humans easily recognize as emotional, e.g. \textit{"Uber and other companies are under fire."}, or \textit{"It's like an Uber for the air. Check out this travel start up."}.

In self-revealing classification, we found two types of recurrent errors, both being false positives. 
The first type stems from the high emotionality of the post being confused with self-revealing, e.g. \textit{"Apple pay sucks."}. 
The second type stems from first-person pronouns, which in those cases do not indicate self-revealing posts, e.g. \textit{"I sent you an email. 
It wouldn't let me post it below."}. 
The last example was classified wrongly in both self-revealing and fact-oriented classifications (as self-revealing and as non fact-oriented). 
In fact-oriented classification, we have mostly seen false negatives, where the models failed to recognize fact-oriented posts, e.g. \textit{"For this phone, battery lasts about 20 minutes but excellent for price."}.
The errors in fact-oriented and self-revealing classification most likely originate from the fact that a great majority of self-revealing posts are at the same time emotional, and a great majority of fact-oriented posts are at the same time non-emotional.
Balancing the two emotionality classes in training datasets for self-revealing and fact-oriented classification task might thus help avoiding some of the errors.

In information-seeking classification, the model was not able to recognize polite information-seeking requests hidden under a layer of modal verbs, e.g. \textit{"I would like to know if anyone would be interested in helping."}. 
The other common types of errors for this model were false positives for the rhetorical questions, such as \textit{"Why can't you help me?"}.

In action-seeking classification, fewer false negatives were found in shorter ($\leq$ 40 words) than in longer posts ($>$ 40words), a 62\% and 75\% out of all misclassified instances, respectively. 
In self-revealing classification, in contrast, we found that longer posts ($>$ 40 words) lead to noticeably fewer false negatives (20\%) than shorter posts (81\%).
\section{Envisioned Usage}

\begin{figure}[!t]
  \centering
  \includegraphics[width=1.0\linewidth]{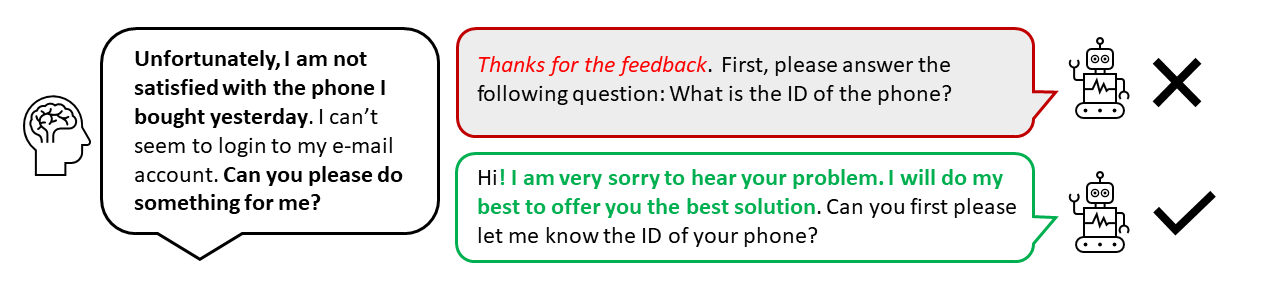}
  \caption{An emotional, self-revealing, and information-seeking utterance from an user.}
  \label{fig:4}
\end{figure} 

\begin{figure}[!t]
  \centering
  \includegraphics[width=1.0\linewidth]{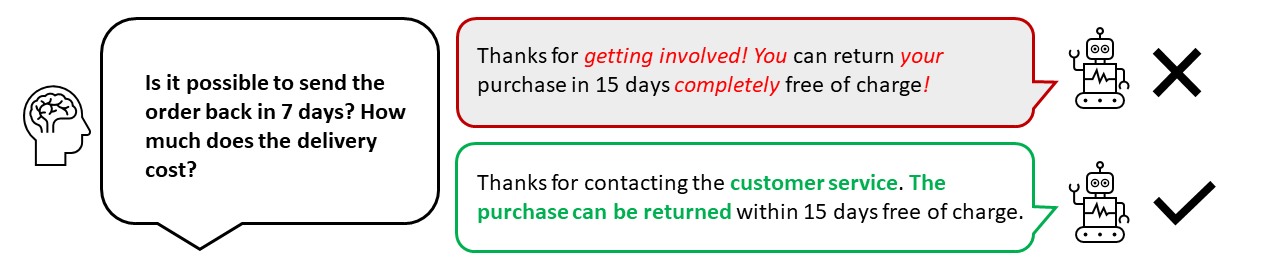}
  \caption{A non-emotional, fact-oriented, action-seeking and information-seeking utterance from an user.}
  \label{fig:5}
\end{figure}

Most commercial chatbots used for customer service and troubleshooting use generic databases to create utterances without resonating with the user's psycholinguistic characteristics and communication needs, and this leads to customers
being unsatisfied and leaving the conversation quickly. 
An effective chatbot should be able to do the same that most humans intuitively do, to adapt to the user's psycholinguistic characteristics, making thus impression of being interested and fully engaged in the conversation with the user.

The works outlined in Sections~\ref{sec:intro} and \ref{sec:method} give pointers to what the right emotionality and communication style would be for each user utterance. 
In terms of emotional tonality, an effective chatbot should respond with equal emotional tonality (emotional or non-emotional). 
When encountered with a user that uses self-revealing communication style (Figure~\ref{fig:4}), chatbot should show adaptive nature, and refer to the user by using second-person pronouns to acknowledge the user's shared personal experience and focus on it. 
When encountered with fact-oriented communication style by a user (Figure~\ref{fig:5}), in contrast, chatbot should keep the wording concise and based on facts. 
In the case of either action-seeking or information-seeking signals from the user's side, if chatbot cannot immediately give the requested answer but rather has to ask something first (Figure~\ref{fig:4}), its answer should contain assurance words such as \textit{recommend} and \textit{offer} giving thus a confirmation that the user's seeking needs will be met in the subsequent answers.

To test those suggestions in a real-world scenario, we obtained 50 real conversations of users with a commercial goal-oriented chatbot, and manually annotated for user satisfaction (\textit{very satisfied}, \textit{neutral}, or \textit{dissatisfied}) based on how the user ended the conversation, considering the following scenarios:

\begin{itemize}
    \item User finishes conversation abruptly and on an angry note (\textit{dissatisfied})
    \item User finishes conversation abruptly without any emotional marker (\textit{neutral})
    \item User finishes conversation on a happy note, or thanking for the conversation (\textit{satisfied})
\end{itemize}

We asked our annotators to label users' psycholinguistic characteristics in those conversations. 
We further annotated the chatbot answers according to the above-mentioned suggestions, for whether it matches users' emotionality and communication styles or not. 
The level of matching between chatbot's and user's psycholinguistic characteristics was calculated as a percentage of psycholinguistic characteristics detected in the user's utterances that were adequately matched by the chatbot's answers. 
Comparison of the user's satisfaction with the level of matching between chatbot's and user's psycholinguistic characteristics confirmed our hypothesis that there is a significant association between the two. 
Although these results can be considered only as preliminary, due to the small sample size and only one chatbot solution taken into account, they give an indication of the potential usage of our framework in the real-world chatbot scenarios for a better user satisfaction.
\section{Conclusion}

In this paper, we proposed five psycholinguistic characteristics that could lead to better written interaction with users in human-to-human interaction, chatbots, and marketing. 
We showed that all five binary classification tasks achieve high performances in five languages (English, Spanish, German, Arabic, and Chinese) using BERT architecture, and that they can be successfully modelled even with small amounts of human-annotated training data (1000 instances).

\section*{Acknowledgements}

We thank Angelo Basile for his support and comments. We also thank all current and past employees of Symanto Research who were involved in earlier stages of this project and thus made this publication possible.

\bibliographystyle{elsarticle-harv}
\bibliography{references}

\end{document}